%% file: neurips_2019.tex
\documentclass{article}

\PassOptionsToPackage{numbers, compress}{natbib}

\usepackage[final]{neurips_2019}
\usepackage{pdfpages}

\newcommand{\appendixpagenumbering}{
  \break
  \pagenumbering{arabic}
  \renewcommand{\thepage}{\arabic{page}}
}
\usepackage[utf8]{inputenc} 
\usepackage[T1]{fontenc}    
\usepackage{hyperref}       
\usepackage{url}            
\usepackage{booktabs}       
\usepackage{amsfonts}       
\usepackage{nicefrac}       
\usepackage{microtype}      
\usepackage{amsmath}
\usepackage{graphicx}
\usepackage{multirow}
\usepackage{multicol}
\usepackage{subfigure}
\usepackage[linesnumbered,ruled]{algorithm2e}
\title{Model-Based Reinforcement Learning with Adversarial Training for Online Recommendation}

\author{%
  Xueying Bai$^{\ast \ddagger}$, Jian Guan\thanks{Both authors contributed equally.} $~^\mathsection$, Hongning Wang$^\dagger$\\
  $^\ddagger$Department of Computer Science, Stony Brook University\\
  $^\mathsection$ Department of Computer Science and Technology, Tsinghua University\\
  $^\dagger$ Department of Computer Science, University of Virginia\\
  \texttt{xubai@cs.stonybrook.edu}, \texttt{j-guan19@mails.tsinghua.edu.cn} \\
  \texttt{hw5x@virginia.edu} 
}

\begin{document}

\maketitle

\begin{abstract}
Reinforcement learning is well suited for optimizing policies of recommender systems. Current solutions mostly focus on model-free approaches, which require frequent interactions with the real environment, and thus are expensive in model learning. Offline evaluation methods, such as importance sampling, can alleviate such limitations, but usually request a large amount of logged data and do not work well when the action space is large. In this work, we propose a model-based reinforcement learning solution which models user-agent interaction for offline policy learning via a generative adversarial network. To reduce bias in the learned model and policy, we use a discriminator to evaluate the quality of generated data and scale the generated rewards. Our theoretical analysis and empirical evaluations demonstrate the effectiveness of our solution in learning policies from the offline and generated data. 
\end{abstract}
\input{introduction}
\input{relwork}
\input{definitions}
\input{method}

\input{theoretical_analy}
\input{experiments}
\input{conclusion}

\bibliographystyle{plainnat}
\bibliography{neurips_2019}
\newpage
\input{Supplementary.tex}
\end{document}

%% file: introduction.tex
\vspace{-0.1cm}
\section{Introduction}
\vspace{-0.1cm}
Recommender systems have been successful in connecting users with their most interested content in a variety of application domains. However, because of users' diverse interest and behavior patterns, only a small fraction of items are presented to each user, with even less feedback recorded. This gives relatively little information on user-system interactions for such a large state and action space~\cite{chen2019top}, and thus brings considerable challenges to construct a useful recommendation policy based on historical interactions. It is important to develop solutions to learn users' preferences from sparse user feedback such as clicks and purchases~\cite{he2016fast,koren2009matrix} to further improve the utility of recommender systems.


Users' interests can be short-term or long-term and reflected by different types of feedback \cite{wu2017returning}. For example, clicks are generally considered as short-term feedback which reflects users' immediate interests during the interaction, while purchase reveals users' long-term interests which usually happen after several clicks. Considering both users' short-term and long-term interests, we frame the recommender system as a reinforcement learning (RL) agent, which aims to maximize users' overall long-term satisfaction without sacrificing the recommendations' short-term utility \cite{shani2005mdp}.     

Classical model-free RL methods require collecting large quantities of data by interacting with the environment, e.g., a population of users. Therefore, without interacting with real users, a recommender cannot easily probe for reward in previously unexplored regions in the state and action space. However, it is prohibitively expensive for a recommender to interact with users for reward and model updates, because bad recommendations (e.g., for exploration) hurt user satisfaction and increase the risk of user drop out. 
In this case, it is preferred for a recommender to learn a policy by fully utilizing the logged data that is acquired from other policies (e.g., previously deployed systems) instead of direct interactions with users.
For this purpose, we take a model-based learning approach in this work, in which we  estimate a model of user behavior from the offline data and use it to interact with our learning agent to obtain an improved policy simultaneously.

Model-based RL has a strong advantage of being sample efficient and helping reduce noise in offline data. However, such an advantage can easily diminish due to the inherent bias in its model approximation of the real environment. Moreover, dramatic changes in subsequent policy updates impose the risk of decreased user satisfaction, i.e., inconsistent recommendations across model updates.  
To address these issues, we introduce adversarial training 
into a recommender's policy learning from offline data. The discriminator is trained to differentiate simulated interaction trajectories from real ones so as to debias the user behavior model and improve policy learning. To the best of our knowledge, this is the first work to explore adversarial training over a model-based RL framework for recommendation. 
We theoretically and empirically demonstrate the value of our proposed solution in policy evaluation. Together, the main contributions of our work are as follows:
\vspace{-0.2cm}
\begin{itemize}
    \item To avoid the high interaction cost, we propose a unified solution to more effectively utilize the logged offline data with model-based RL algorithms, integrated via adversarial training. It enables robust recommendation policy learning.
    \item The proposed model is verified through theoretical analysis and extensive empirical evaluations. Experiment results demonstrate our solution's better sample efficiency over the state-of-the-art baselines \footnote{Our implementation is available at \url{https://github.com/JianGuanTHU/IRecGAN}.} 
\end{itemize}

%% file: relwork.tex
\vspace{-0.1cm}
\section{Related Work}
\vspace{-0.1cm}
\textbf{Deep RL for recommendation} 
There have been studies utilizing deep RL solutions in news, music and video recommendations \cite{lu2016partially,liebman2015dj,zheng2018drn}. However, most of the existing solutions are model-free methods and thus do not explicitly model the agent-user interactions. In these methods, value-based approaches, such as deep Q-learning \cite{mnih2015human}, present unique advantages such as seamless off-policy learning, but are prone to instability with function approximation \cite{sutton2000policy,mnih2013playing}. And the policy's convergence in these algorithms is not well-studied. In contrast, policy-based methods such as policy gradient \cite{learning1998introduction} remain stable but suffer from data bias without real-time interactive control due to learning and infrastructure constraints. Oftentimes, importance sampling \cite{munos2016safe} is adopted to address the bias but instead results in huge variance \cite{chen2019top}. In this work, we rely on a policy gradient based RL approach, in particular, REINFORCE \cite{williams1992simple}; but we simultaneously estimate a user behavior model to provide a reliable environment estimate so as to update our agent on policy.

\textbf{Model-based RL}
 Model-based RL algorithms incorporate a model of the environment to predict rewards for unseen state-action pairs. It is known in general to outperform model-free solutions in terms of sample complexity \cite{deisenroth2013survey}, and has been applied successfully to control robotic systems both in simulation and real world \cite{deisenroth2011pilco,meger2015learning,morimoto2003minimax,deisenroth2011learning}. Furthermore, Dyna-Q \cite{sutton1990integrated,peng2018deep} integrates model-free and model-based RL to generate samples for learning in addition to the real interaction data. \citet{gu2016continuous} extended these ideas to neural network models, and \citet{peng2018deep} further apply the method on task-completion dialogue policy learning. However, the most efficient model-based algorithms have used relatively simple function approximations, 
 which actually have difficulties in high-dimensional space with nonlinear dynamics and thus lead to huge approximation bias. 


\textbf{Offline evaluation} The problems of off-policy learning  \cite{munos2016safe,precup2000eligibility,precup2001off} and offline policy evaluation are generally pervasive and challenging in RL, and in recommender systems in particular. As a policy evolves, so does the distribution under which the expectation of gradient is computed. Especially in the scenario of recommender systems, where item catalogues and user behavior change rapidly, substantial policy changes are required; and therefore it is not feasible to take the classic approaches \cite{schulman2015trust,achiam2017constrained} to constrain the policy updates before new data is collected under an updated policy. Multiple off-policy estimators leveraging inverse-propensity scores, capped inverse-propensity scores and various variance control measures have been developed \cite{thomas2016data,swaminathan2015self,swaminathan2015batch,gilotte2018offline} for this purpose. 

\textbf{RL with adversarial training}
\citet{yu2017seqgan} propose SeqGAN to extend GANs with an RL-like generator for the sequence generation problem, where the reward signal is provided by the discriminator at the end of each episode via a Monte Carlo sampling approach. The generator takes sequential actions and learns the policy using estimated cumulative rewards. In our solution, the generator consists of two components, i.e., our recommendation agent and the user behavior model, and we model the interactive process via adversarial training and policy gradient. Different from the sequence generation task which only aims to generate sequences similar to the given observations, we leverage adversarial training to help reduce bias in the user model and further reduce the variance in training our agent. The agent learns from both the interactions with the user behavior model and those stored in the logged offline data. To the best of our knowledge, this is the first work that utilizes adversarial training for improving both model approximation and policy learning on offline data.

%% file: definitions.tex
\vspace{-0.1cm}
\section{Problem Statement}
\vspace{-0.1cm}
The problem is to learn a policy from offline data such that when deployed online it maximizes cumulative rewards collected from interactions with users. We address this problem with a model-based reinforcement learning solution, which explicitly model users' behavior patterns from data. 

\textbf{Problem}
A recommender is formed as a learning agent to generate actions under a policy, where each action gives a recommendation list of $k$ items. Every time through interactions between the agent and the environment (i.e., users of the system), a set $\Omega$ of $n$ sequences $\Omega = \{\tau_1,...,\tau_{n}\}$ is recorded, where $\tau_i$ is the $i$-{th} sequence containing agent actions, user behaviors and rewards: $\tau_i = \{(a_0^i, c_0^i, r_0^i), (a_1^i, c_1^i, r_1^i),...,(a_{t}^i, c_t^i, r_{t}^i)\}$, $r_t^i$ represents the reward on $a_t^i$ (e.g., make a purchase), and $c_t^i$ is the associated user behavior corresponding to agent's action $a_t^i$ (e.g., click on a recommended item). 
For simplicity, in the rest of paper, we drop the superscript $i$ to represent a general sequence $\tau$. Based on the observed sequences, a policy $\pi$ is learnt to maximize the expected cumulative reward $\mathbb{E}_{\tau\sim\pi}[\sum_{t=0}^{T}r_t]$, where $T$ is the end time of $\tau$. 

\textbf{Assumption}
To narrow the scope of our discussion, we study a typical type of user behavior, i.e., clicks, and make following assumptions: 1) at each time a user must click on one item from the recommendation list; 2) items not clicked in the recommendation list will not influence the user's future behaviors; 3) rewards only relate to clicked items. For example, when taking the user's purchase as reward, purchases can only happen in the clicked items.  

\textbf{Learning framework}
In a Markov Decision Process, an environment consists of a state set $S$, an action set $A$, a state transition distribution $P: S \times A \times S$, and a reward function $f_r: S\times A \to \mathbb{R}$, which maps a state-action pair to a real-valued scalar. 
In this paper, the environment is modeled as a user behavior model $\mathcal{U}$, and learnt from offline log data. $S$ is reflected by the interaction history before time $t$, and $P$ captures the transition of user behaviors. In the meanwhile, based on the assumptions mentioned above, 
at each time $t$, the environment generates user's click $c_{t}$ on items recommended by an agent $\mathcal{A}$ in $a_t$ based on his/her click probabilities under the current state; and the reward function $f_r$ generates reward $r_{t}$ for the clicked item $c_t$. 

\begin{figure}[!ht]
\centering
\includegraphics[width=5.5in]{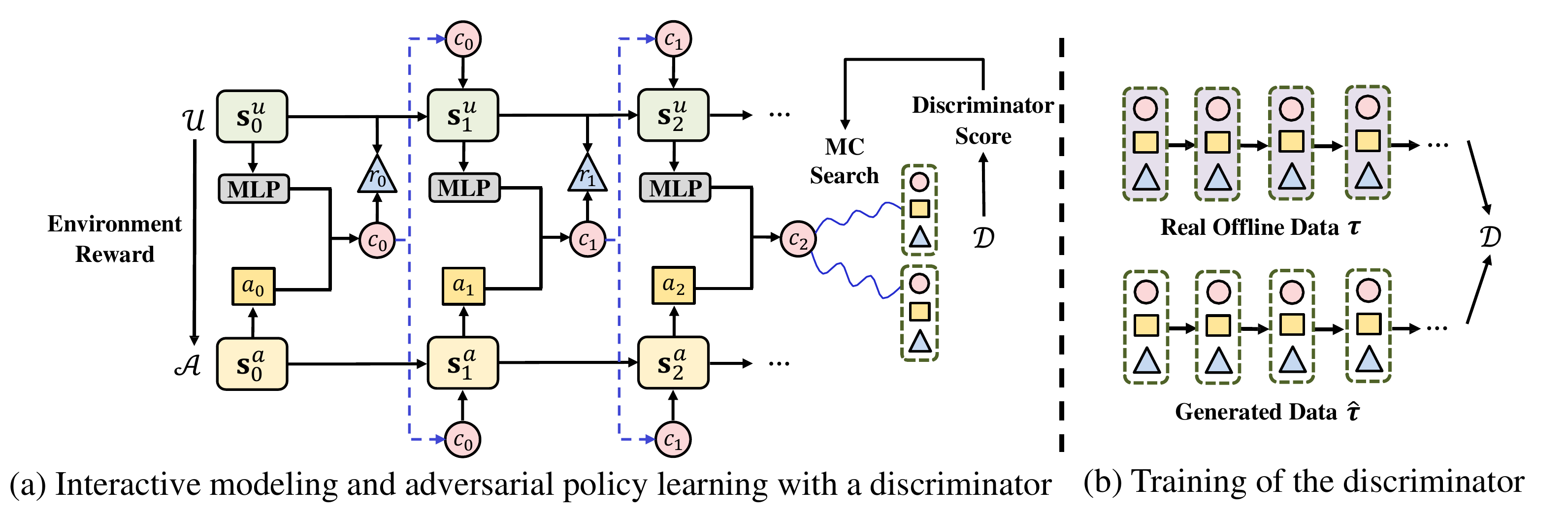} 
\caption{Model overview of IRecGAN. $\mathcal{A}$, $\mathcal{U}$ and $\mathcal{D}$ denote the agent model, user behavior model, and discriminator, respectively. In IRecGAN, $\mathcal{A}$ and $\mathcal{U}$ interact with each other to generate recommendation sequences that are close to the true data distribution, so as to jointly reduce bias in $\mathcal{U}$ and improve the recommendation quality in $\mathcal{A}$.}
\label{fig:model_overview}
\end{figure}

Our recommendation policy is learnt from both offline data and data sampled from the learnt user behavior model, i.e., a model-based RL solution.
We incorporate adversarial training in our model-based policy learning to: 1) improve the user model to ensure the sampled data is close to true data distribution; 
2) utilize the discriminator to scale rewards from generated sequences to further reduce bias in value estimation. Our proposed solution contains an interactive model constructed by $\cal U$ and $\cal A$, and an adversarial policy learning approach. We name the solution as Interactive Recommender GAN, or IRecGAN in short. The overview of our proposed solution is shown in Figure \ref{fig:model_overview}.

%% file: method.tex
\vspace{-0.1cm}
\section{Interactive Modeling for Recommendation}
\vspace{-0.1cm}
We present our interactive model for recommendation, which consists of two components: 1) the user behavior model $\mathcal{U}$ that generates user clicks over the recommended items with corresponding rewards; and 2) the agent $\mathcal{A}$ which generates recommendations according to its policy. $\mathcal{U}$ and $\mathcal{A}$ interact with each other to generate user behavior sequences for adversarial policy learning. 

\textbf{User behavior model}
Given users' click observations $\{c_0, c_1,...,c_{t-1}\}$, the user behavior model $\mathcal{U}$ first projects the clicked item into an embedding vector $\mathbf{e}^u$ at each time \footnote{As we can use different embeddings on the user side and agent side, we use the superscript $u$ and $a$ to denote this difference accordingly.}. The state $\mathbf{s}^u_t$ can be represented as a summary of click history, i.e., $\mathbf{s}^u_t = h_u(\mathbf{e}^u_0, \mathbf{e}^u_1, ...\mathbf{e}^u_{t-1})$. We use a recurrent neural network to model the state transition $P$ on the user side, thus for the state $\mathbf{s}^u_t$ we have,
\begin{equation*}
    \mathbf{s}^u_t = h^u(\mathbf{s}^u_{t-1}, \mathbf{e}^u_{t-1}),
\end{equation*}
$h^u(\cdot, \cdot)$ can be functions in the RNN family like GRU \cite{chung2014empirical} and LSTM \cite{hochreiter1997long} cells. Given the action $a_t=\{a_{t(1)},...a_{t(k)}\}$, i.e., the top-$k$ recommendations at time $t$, we compute the probability of click among the recommended items via a softmax function, 
\begin{equation}
    {\mathbf{V}^c} = ({\mathbf{W}^c}{\mathbf{s}^u_t} + {\mathbf{b}^c})^\top\mathbf{E}_t^u , ~~
    p(c_{t}|\mathbf{s}^u_t, a_t) = \exp(\mathbf{V}^c_i)/\sum\nolimits^{|a_t|}_{j=1} {\exp(\mathbf{V}^c_j)}
    \label{user_prob}
\end{equation}
where $\mathbf{V}^c\in \mathbb{R}^k$ is a transformed vector indicating the evaluated quality of each recommended item $a_{t(i)}$ under state $\mathbf{s}^u_t$, ${\mathbf{E}^u_t}$ is the embedding matrix of recommended items, $\mathbf{W}^c$ is the click weight matrix, and $\mathbf{b}^c$ is the corresponding bias term.
Under the assumption that target rewards only relate to clicked items, the reward $r_t$ for ($\mathbf{s}^u_t$, $a_t$) is calculated by:
\begin{equation}
    r_t(\mathbf{s}^u_t, a_{t}) = f_r\big(({\mathbf{W}^r}{\mathbf{s}^u_t} + {\mathbf{b}^r})^\top \mathbf{e}^u_{t}\big),
    \label{eq_reward}
\end{equation}
where $\mathbf{W}^r$ is the reward weight matrix, $\mathbf{b}^r$ is the corresponding bias term, and $f_r$ is the reward mapping function and can be set according to the reward definition in specific recommender systems. For example, if we make $r_t$ the purchase of a clicked item $c_t$, where $r_t=1$ if it is purchased and $r_t=0$ otherwise, $f_r$ can be realized by a Sigmoid function with binary output.

Based on Eq \eqref{user_prob} and \eqref{eq_reward}, taking categorical reward, the user behavior model $\mathcal{U}$ can be estimated from the offline data $\Omega$ via maximum likelihood estimation:
\begin{equation}
    \max \sum_{\tau_i\in\Omega}\sum\nolimits_{t = 0}^{T_i} {\log p({c^i_{t}}|\mathbf{s}^{u_i}_t, a^i_t) + \lambda_p \log p({r^i_t}|\mathbf{s}^{u_i}_t,{c^i_{t}})},  \label{U_pretrain}
\end{equation}
where $\lambda_p$ is a parameter balancing the loss between click prediction and reward prediction, and $T_i$ is the length of the observation sequence $\tau_i$. With a learnt user behavior model, user clicks and reward on the recommendation list can be sampled from Eq \eqref{user_prob} and \eqref{eq_reward} accordingly.

\textbf{Agent} The agent should take actions based on the environment's provided states. However, in practice, users' states are not observable in a recommender system. Besides, as discussed in \cite{oh2017value}, the states for the agent to take actions may be different from those for users to generate clicks and rewards.  
As a result, we build a different state model on the agent side in $\mathcal{A}$ to learn its states. 
Similar to that on the user side, given the projected click vectors $\{\mathbf{e}^a_0, \mathbf{e}^a_2,...\mathbf{e}^a_{t-1}\}$, we model states on the agent side by $\mathbf{s}^a_t = h^a(\mathbf{s}^a_{t-1}, \mathbf{e}^a_{t-1})$, where $\mathbf{s}_t^a$ denotes the state maintained by the agent at time $t$, $h^a(\cdot, \cdot)$ is the chosen RNN cell. The initial state $\mathbf{s}_0^a$ for the first recommendation is drawn from a distribution $\rho$. We simply denote it as $\mathbf{s}_0$ in the rest of our paper.  
We should note that although the agent also models states based on users' click history, it might create different state sequences than those on the user side. 

Based on the current state $\mathbf{s}_t^a$, the agent generates a size-$k$ recommendation list out of the entire set of items as its action $a_t$. The probability of item $i$ to be included in $a_t$ under the policy $\pi$ is:
\begin{equation}
    \pi(i \in {a_t}|{\mathbf{s}^a_t}) =  \frac{{\exp (\mathbf{W}^a_i\mathbf{s}^a_t + \mathbf{b}^a_i)}}{{\sum\nolimits_{j = 1}^{|C|} {\exp (\mathbf{W}^a_j\mathbf{s}^a_t + \mathbf{b}^a_j)} }}, \label{agent_prob}
\end{equation}
where $\mathbf{W}_i^a$ is the $i$-th row of the action weight matrix $\mathbf{W}^a$, $C$ is the entire set of recommendation candidates, and $\mathbf{b}_i^a$ is the corresponding bias term. Following \cite{chen2019top}, we generate $a_t$ by sampling without replacement according to Eq (\ref{agent_prob}). Unlike \cite{pmlr-v97-chen19f}, we do not consider the combinatorial effect among the $k$ items by simply assuming the users will evaluate them independently (as indicated in Eq \eqref{user_prob}).

\vspace{-0.1cm}
\section{Adversarial Policy Learning}
\vspace{-0.1cm}
We use the policy gradient method REINFORCE \cite{williams1992simple} for the agent's policy learning, based on both the generated and offline data.
When generating $\hat\tau_{0:t} = \big\{(\hat{a}_0, \hat{c}_0, \hat{r}_0),...,(\hat{a}_{t}, \hat{c}_t, \hat{r}_{t})\big\}$ for $t>0$, 
we obtain $\hat{a}_{t}=\mathcal{A}(\hat \tau^c_{0:t-1})$ by Eq \eqref{agent_prob}, $\hat{c}_{t}=\mathcal{U}_c(\hat \tau^c_{0:t-1}, \hat{a}_{t})$ by Eq \eqref{eq_reward}, and $\hat{r}_{t}=\mathcal{U}_r(\hat \tau^c_{0:t-1},\hat{c}_{t})$ by Eq \eqref{user_prob}. $\tau^c$ represents clicks in the sequence $\tau$ and $(\hat{a}_0, \hat{c}_0, \hat{r}_0)$ is generated by $\mathbf{s}_0^a$ and $\mathbf{s}_0^u$ accordingly.
The generation of a sequence ends at the time $t$ if $\hat c_t=c_{end}$, where $c_{end}$ is a stopping symbol. The distributions of generated and offline data are denoted as $g$ and $data$ respectively. In the following discussions, we do not explicitly differentiate $\tau$ and $\hat \tau$ when the distribution of them is specified.
Since we start the training of $\mathcal{U}$ from offline data, it introduces inherent bias from the observations and our specific modeling choices. The bias affects the sequence generation and thus may cause biased value estimation. To reduce the effect of bias, we apply adversarial training to control the training of 
both $\mathcal{U}$ and $\mathcal{A}$. The discriminator is also used to rescale the generated rewards $\hat r$ for policy learning. Therefore, the learning of agent $\cal A$ considers both sequence generation and target rewards.
 
\subsection{Adversarial training}
We leverage adversarial training to encourage our IRecGAN model to generate high-quality sequences that capture intrinsic patterns in the real data distribution. A discriminator $\mathcal{D}$ is used to evaluate a given sequence $\tau$, where $\mathcal{D}(\tau)$ represents the probability that $\tau$ is generated from the real recommendation environment. 
The discriminator can be estimated by minimizing the objective function:
\begin{equation}
    - {{\mathbb E}_{\tau  \sim {data}}}\log \big(\mathcal{D}(\tau )\big) - {{\mathbb E}_{\tau  \sim g}}\log \big(1 - \mathcal{D}(\tau )\big) \label{D_train}.
\end{equation}
However, $\mathcal{D}$ only evaluates a completed sequence, and hence it cannot directly evaluate a partially generated sequence at a particular time step $t$. Inspired by \cite{yu2017seqgan}, we utilize the Monte-Carlo tree search algorithm with the roll-out policy constructed by $\mathcal{U}$ and $\mathcal{A}$ to get sequence generation score at each time.
%
At time $t$, the sequence generation score $q_\mathcal{D}$ of $\tau_{0:t}$ is defined as:
\begin{equation}
    q_{\cal D}(\tau_{0:t}) =  \left\{ {\begin{array}{*{20}{l}}
{\frac{1}{N}\sum\nolimits_{n = 1}^N {\mathcal{D}(\tau _{_{0:T}}^n)} ,\tau _{_{0:T}}^n \in MC^{\mathcal{U},\mathcal{A}}({\tau _{_{0:t}}};N)}&{t < T}\\
{\mathcal{D}({\tau _{_{0:T}}})}&{t = T}
\end{array}} \right. \label{roll-out}
\end{equation}
where $MC^{\mathcal{U},\mathcal{A}}({\tau _{_{0:t}}};N)$ is the set of $N$ sequences sampled from the interaction between $\mathcal{U}$ and $\mathcal{A}$. 

Given the observations in offline data, $\mathcal{U}$ should generate clicks and rewards that reflect intrinsic patterns of the real data distribution. Therefore, $\mathcal{U}$ should maximize the sequence generation objective ${\mathbb{E}}_{{\mathbf{s}_0^u \sim \rho}}[\sum\nolimits_{(a_0, c_0, r_0) \sim g}{{\mathcal{U}}({c_0, r_0}|\mathbf{s}_0^u, a_0)}  \cdot q_{\cal{D}}({\tau_{0:0}})]$, which is the expected discriminator score for generating a sequence from the initial state. $\mathcal{U}$ may not generate clicks and rewards exactly the same as those in offline data, but the similarity of its generated data to offline data is still an informative signal to evaluate its sequence generation quality. 
By setting $q_{\cal{D}}(\tau_{0:t})=1$ at any time $t$ for offline data, we extend this objective to include offline data (it becomes the data likelihood function on offline data). Following \cite{yu2017seqgan}, based on Eq \eqref{user_prob} and Eq \eqref{eq_reward}, the gradient of $\mathcal{U}$'s objective can be derived as,
\begin{equation}
    {{\mathbb E}_{\tau \sim {\{{g,data}\}}}}\Big[\sum\nolimits_{t = 0}^T {q_{\cal{D}}({\tau _{0:t}})} {\nabla _{\Theta_u}}\big( \log p_{\Theta_u}({c}_{t}|\mathbf{s}^u_t, {a}_t) + \lambda_p \log p_{\Theta_u}({r_t}|\mathbf{s}^{u}_t,{c_{t}})\big)\Big],
    \label{seq_usr}
\end{equation}
where $\Theta_u$ denotes the parameters of $\mathcal{U}$ and $\Theta_a$ denotes those of $\mathcal{A}$. 
Based on our assumption, even when $\mathcal{U}$ can already capture users' true behavior patterns, it still depends on $\mathcal{A}$ to provide appropriate recommendations to generate clicks and rewards that the discriminator will treat as authentic. Hence, $\mathcal{A}$ and $\mathcal{U}$ are coupled in this adversarial training. To encourage $\mathcal{A}$ to provide \textit{needed} recommendations, we include $q_{\cal{D}}(\tau_{0:t})$ as a sequence generation reward for $\mathcal{A}$ at time $t$ as well. As $q_{\cal{D}}(\tau_{0:t})$ evaluates the overall generation quality of $\tau_{0:t}$, it ignores sequence generations after $t$. To evaluate the quality of a whole sequence, we require $\mathcal{A}$ to maximize the cumulative sequence generation reward ${\mathbb{E}_{\tau \sim \{g, data\}}}\big[\sum\nolimits_{t = 0}^T {q_\mathcal{D}({\tau _{0:t}})}\big ] $. Because $\mathcal{A}$ does not directly generate the observations in the interaction sequence, we approximate ${\nabla _{{\Theta _{a}}}}\big (\sum\nolimits_{t = 0}^T {{q_{\cal D}}({\tau _{0:t}})}\big)$ as 0 when calculating the gradients. Putting these together, the gradient derived from sequence generations for $\mathcal{A}$ is estimated as, 
\begin{equation}
    {{\mathbb E}_{\tau \sim \{ g,data\} }}\big[\sum\nolimits_{t = 0}^T {\big(\sum\nolimits_{t' = t}^T {\gamma ^{t'-t}{q_{\cal{D}}}({\tau _{0:t}})} \big)} {\nabla _{{\Theta _{a}}}}\log {\pi _{{\Theta _{a}}}}({c_t} \in {a_t}|{\bf{s}}_t^a)\big].
    \label{seq_agent}
\end{equation}
Based on our assumption that only the clicked items influence user behaviors, and $\mathcal{U}$ only generates rewards on the clicked items, we use ${\pi _{\Theta_a} }({c_{t} \in a_t}|s_t^a)$ as an estimation of ${\pi _{\Theta_a} }({a_t}|s_t^a)$, i.e., $\mathcal{A}$ should promote $c_t$ in its recommendation at time $t$. In practice, we add a discount factor $\gamma<1$ when calculating the cumulative rewards to reduce estimation variance \cite{chen2019top}.

\subsection{Policy learning}
Because our adversarial training encourages IRecGAN to generate clicks and rewards with similar patterns as offline data, and we assume rewards only relate to the clicked items, we use offline data as well as generated data for policy learning.
Given data $\tau_{0:T}=\{(a_0, c_0, r_0),...,(a_{T}, c_T, r_{T})\}$, including both offline and generated data, the objective of the agent is to maximize the expected cumulative reward ${\mathbb{E}_{\tau \sim {\{{g,data}\}}}}[{R_T}]$, where ${R_T} = \sum\nolimits_{t = 0}^{T} {{r_t}}$. In the generated data, due to the difference in distributions of the generated and offline sequences, the generated reward $\hat{r}_t$ calculated by Eq \eqref{eq_reward} might be biased. To reduce such bias, we utilize the sequence generation score in Eq \eqref{roll-out} to rescale the generated rewards: $r^s_t = q_{\cal D}(\tau_{0:t}) \hat{r}_t$, and treat it as the reward for generated data. The gradient of the objective is thus estimated by:
\begin{equation}
    {\mathbb{E}_{\tau \sim {\{{g,data}\}}}}\big[\sum\nolimits_{t=0}^{T} {{R_t}{\nabla _{\Theta_a} }\log {\pi _{\Theta_a} }({c_{t} \in a_t}|\mathbf{s}_t^a)} \big],~~ {R_t} = \sum\nolimits_{t' = t}^{T} {\gamma ^{t' - t}} q_{\cal D}(\tau_{0:t}){r_{t}}
    \label{pur_agent}
\end{equation}

$R_t$ is an approximation of $R_T$ with the discount factor $\gamma$. Overall, the user behavior model ${\mathcal{U}}$ is updated only by the sequence generation objective defined in Eq \eqref{seq_usr} on both offline and generated data; but the agent ${\mathcal{A}}$ is updated by both sequence generation and target rewards. 
Hence, the overall reward for ${\mathcal{A}}$ at time $t$ is 
${q_{\cal D}({\tau _{0:t}})(1 +  \lambda_r{r_t})}$, where $\lambda_r$ is the weight for cumulative target rewards. The overall gradient for $\mathcal{A}$ is thus: 
\begin{equation}
      {\mathbb{E}_{\tau \sim \{ g,data\} }}\big[\sum\nolimits_{t = 0}^T R^a_t{{\nabla _{{\Theta _a}}}\log {\pi _{{\Theta _a}}}({c_{t}} \in {a_t}|\mathbf{s}_t^a)}\big], ~~R^a_t= {\sum\nolimits_{t' = t}^T {{\gamma ^{t' - t}}} q_{\cal D}({\tau _{0:t}})(1 + {\lambda _r}{r_t})} 
    \label{all_agent}
\end{equation}

%% file: theoretical_analy.tex
\vspace{-0.1cm}
\section{Theoretical Analysis}
\vspace{-0.1cm}
For one iteration of policy learning in IRecGAN, we first train the discriminator $\mathcal{D}$ with offline data, which follows $P_{data}$ and was generated by an unknown logging policy, and the data generated by IRecGAN under $\pi_{\Theta_a}$ with the distribution of $g$. When $\Theta_u$ and $\Theta_a$ are learnt, for a given  sequence $\tau$, by \textit{proposition} 1 in \cite{goodfellow2014generative}, the optimal discriminator $\mathcal{D}$ is $\mathcal{D}^*(\tau)=\frac{{{P_{{{data}}}}(\tau)}}{{{P_{{{data}}}}(\tau) + {P_g}(\tau)}}$.  

\textbf{Sequence generation}
Both $\mathcal{A}$ and $\mathcal{U}$ contribute to the sequence generation in IRecGAN. $\mathcal{U}$ is updated by the gradient in Eq \eqref{seq_usr} to maximize the sequence generation objective. At time $t$, the expected sequence generation reward for $\mathcal{A}$ on the generated data is: 
$
{E_{\tau_{0:t} \sim g}[q_\mathcal{D}({\tau _{0:t}})] 
 = E_{\tau_{0:t} \sim g}[\mathcal{D}(\tau _{0:T}|{\tau _{0:t}})}].
$
The expected value on $\tau _{0:t}$ is: 
$
{\mathbb{E}_{\tau \sim g}}[V_g] = {\mathbb{E}_{\tau \sim g} }\big[\sum\nolimits_{t = 0}^T {q_\mathcal{D}({\tau _{0:t}})}\big ] = \sum\nolimits_{t = 0}^T {{\mathbb{E}_{{\tau _{0:t}}\sim g}}\big[\mathcal{D}({\tau _{0:T}}|{\tau _{0:t}})\big]}. 
\label{sequence_gen}
$
Given the optimal $\mathcal{D}^*$, the sequence generation value can be written as:
\begin{equation}
    {\mathbb{E}_{\tau\sim g} }[V_g^{}] = \sum\nolimits_{t = 0}^T {{\mathbb{E}_{{\tau _{0:t}}\sim g}}\Big[\frac{{{P_{data}}({\tau _{0:T}}|{\tau _{0:t}})}}{{{P_{data}}({\tau _{0:T}}|{\tau _{0:t}}) + {P_g}({\tau _{0:T}}|{\tau _{0:t}})}}\Big]}.
    \label{Gan_obj}
\end{equation}
Maximizing each term in the summation of Eq \eqref{Gan_obj} is an objective for the generator at time $t$ in GAN. According to \cite{goodfellow2014generative}, the optimal solution for all such terms is ${P_g(\tau_{0:T}|s_0)} = {P_{{{data}}}(\tau_{0:T}|s_0)}$. It means $\cal A$ can maximize the sequence generation value when it helps to generate sequences with the same distribution as $data$. Besides the global optimal, Eq \eqref{Gan_obj} also encourages $\mathcal{A}$ to reward each ${P_g(\tau_{0:T}|\tau_{0:t})} = {P_{{{data}}}(\tau_{0:T}|\tau_{0:t})}$, even if $\tau_{0:t}$ is less likely to be generated from $P_g$. 
This prevents IRecGAN to recommend items only considering users' immediate preferences.

\textbf{Value estimation} The agent $\mathcal{A}$ should also be updated to maximize the expected value of target rewards $V_a$. To achieve this, we use discriminator $\mathcal{D}$ to rescale the estimation of $V_a$ on the generated sequences, and we also combine offline data to evaluate $V_a$ for policy $\pi_{\Theta_a}$:
\begin{equation}
    {{\mathbb E}_{\tau  \sim {\pi _{_{{\Theta _a}}}}}}[V_a] = {\lambda _1}\sum\nolimits_{t = 0}^T {{{\mathbb E}_{{\tau _{0:t}} \sim g}}\frac{{{P_{data}}(\tau _{0:t}^{})}}{{{P_{data}}(\tau _{0:t}^{}) + {P_g}(\tau _{0:t}^{})}}} {{\hat r}_t} + {\lambda _2}\sum\nolimits_{t = 0}^T {{{\mathbb E}_{{\tau _{0:t}} \sim data}}} {r_t},
    \label{expectation}
\end{equation}
where ${{{\hat r}_t}}$ is the generated reward by $\mathcal{U}$ at time $t$ and $r_t$ is the true reward. $\lambda_1$ and $\lambda_2$ represent the ratio of generated data and offline data during model training, and we require $\lambda_1 + \lambda_2 = 1$. Here we simplify $P(\tau_{0:T}|\tau_{0:t})$ as $P(\tau_{0:t})$. As a result, there are three sources of biases in this value estimation:
\[\Delta  = {{\hat r}_t} - {r_t},\begin{array}{*{20}{c}}
{}
\end{array}{\delta _1} = 1 -{{{P_{{\pi _{{\Theta_a}}}}}(\tau _{0:t}^{})}}/{{{P_g}(\tau _{0:t}^{})}},\begin{array}{*{20}{c}}
{}
\end{array}{\delta _2} = 1 - {{{P_{{\pi _{{\Theta_a}}}}}(\tau _{0:t}^{})}}/{{{P_{data}}(\tau _{0:t}^{})}}.\]
Based on different sources of biases, the expected value estimation in Eq \eqref{expectation} is:
\small
\begin{align*}
{{\mathbb E}_{\tau  \sim {\pi _{_{{\Theta_a}}}}}}[V_a] = &{\lambda _1}\sum\limits_{t = 0}^T {{{\mathbb E}_{{\tau _{0:t}} \sim g}}\frac{{{P_{{\pi _{{\Theta _a}}}}}(\tau _{0:t}^{})}}{{{P_g}(\tau _{0:t}^{})}}\frac{\Delta  + {r_t}}{{2 - ({\delta _1} + {\delta _2})}}}+ {\lambda _2}\sum\limits_{t = 0}^T {{{\mathbb E}_{{\tau _{0:t}} \sim data}}\Big(\frac{{{P_{{\pi _{{\Theta_a}}}}}(\tau _{0:t}^{})}}{{{P_{data}}(\tau _{0:t}^{})}} + {\delta_2}\Big)} {r_t}\\
= & V_a^{{\pi _{{\Theta _a}}}} + \sum\limits_{t = 0}^T {{{\mathbb E}_{{\tau _{0:t}}\sim{\pi _{{\Theta _a}}}}}{w_t}} \Delta  + \sum\limits_{t = 0}^T {{{\mathbb E}_{{\tau _{0:t}}\sim data}}{\lambda _2}{\delta _2}{r_t}} - \sum\limits_{t = 0}^T {{{\mathbb E}_{{\tau _{0:t}}\sim{\pi _{{\Theta _a}}}}}} ({\lambda _1} - {w_t}){r_t},
\end{align*}
\normalsize
where ${w_t} = \frac{{{\lambda _1}}}{{2 - ({\delta _1} + {\delta _2})}}$. 
$\Delta$ and $\delta_1$ come from the bias of user behavior model $\cal U$. Because the adversarial training helps  improve $\cal U$ to capture real data patterns, it decreases $\Delta$ and $\delta_2$. Because we can adjust the sampling ratio $\lambda_1$ to reduce $w_t$, $w_t\Delta$ can be small. The sequence generation rewards for agent $\mathcal A$ encourage distribution $g$ to be close to ${data}$. Because ${\delta _2} = 1 - \frac{{{P_{{\pi _{{\Theta _a}}}}}({\tau _{0:t}})}}{{{P_g}({\tau _{0:t}})}} \cdot \frac{{{P_g}({\tau _{0:t}})}}{{{P_{data}}({\tau _{0:t}})}}$, the bias $\delta_2$ can also be reduced. It shows our method has a bias controlling effect. 


%% file: experiments.tex
\vspace{-0.1cm}
\section{Experiments}
\vspace{-0.1cm}
In our theoretical analysis, we can find that reducing the model bias improves value estimation, and therefore improves policy learning. In this section, we conduct empirical evaluations on both real-world and synthetic datasets to demonstrate that our solution can effectively model the pattern of data for better recommendations, compared with state-of-the-art solutions.\\

\vspace{-0.5cm}
\subsection{Simulated Online Test}

Subject to the difficulty of deploying a recommender system with real users for online evaluation, we use simulation-based studies to first investigate the effectiveness of our approach following \cite{zhao2019model,pmlr-v97-chen19f}. 

\textbf{Simulated Environment} 
We synthesize an MDP to simulate an online recommendation environment. It has $m$ states and $n$ items for recommendation, with a randomly initialized transition probability matrix $P(s\in S|a_j\in A, s_i\in S)$. Under each state $s_i$, an item $a_j$'s reward $r(a_j\in A|s_i\in S)$ is uniformly sampled from the range of 0 to 1. During the interaction, given a recommendation list including $k$ items selected from the whole item set by an agent, the simulator first samples an item proportional to its ground-truth reward under the current state $s_i$ as the click candidate. Denote the sampled item as $a_j$, a Bernoulli experiment is performed on this item with $r(a_j)$ as the success probability; then the simulator moves to the next state according to the state transition probability $p(s|a_j,s_i)$. A special state $s_0$ is used to initialize all the sessions, which do not stop until the Bernoulli experiment fails. The immediate reward is 1 if the session continues to the next step; otherwise 0. 
In our experiment, $m,n$ and $k$ are set to 10, 50 and 10 respectively. 


\textbf{Offline Data Generation} 
We generate offline recommendation logs denoted by $d_{\text{off}}$ with the simulator. The bias and variance in $d_{\text{off}}$ are especially controlled by changing the logging policy and the size of $d_{\text{off}}$. 
We adopt three different logging policies: 1) uniformly random policy $\pi_{\text{random}}$, 2) maximum reward policy $\pi_{\text{max}}$, 3) mixed reward policy $\pi_{\text{mix}}$. Specifically, $\pi_{\text{max}}$ recommends the top $k$  items with the highest ground-truth reward under the current simulator state at each step, while $\pi_{\text{mix}}$ randomly selects $k$ items with either the top 20\%-50\% ground-truth reward or the highest ground-truth reward under a given state. In the meanwhile, we vary the size of data in $d_{\text{off}}$ from 200 to 10,000.

\begin{figure}[t]
\vspace{-1mm}
\centering
\subfigure[$\pi_{\text{random}}(200)$]{\includegraphics[width=2.73in]{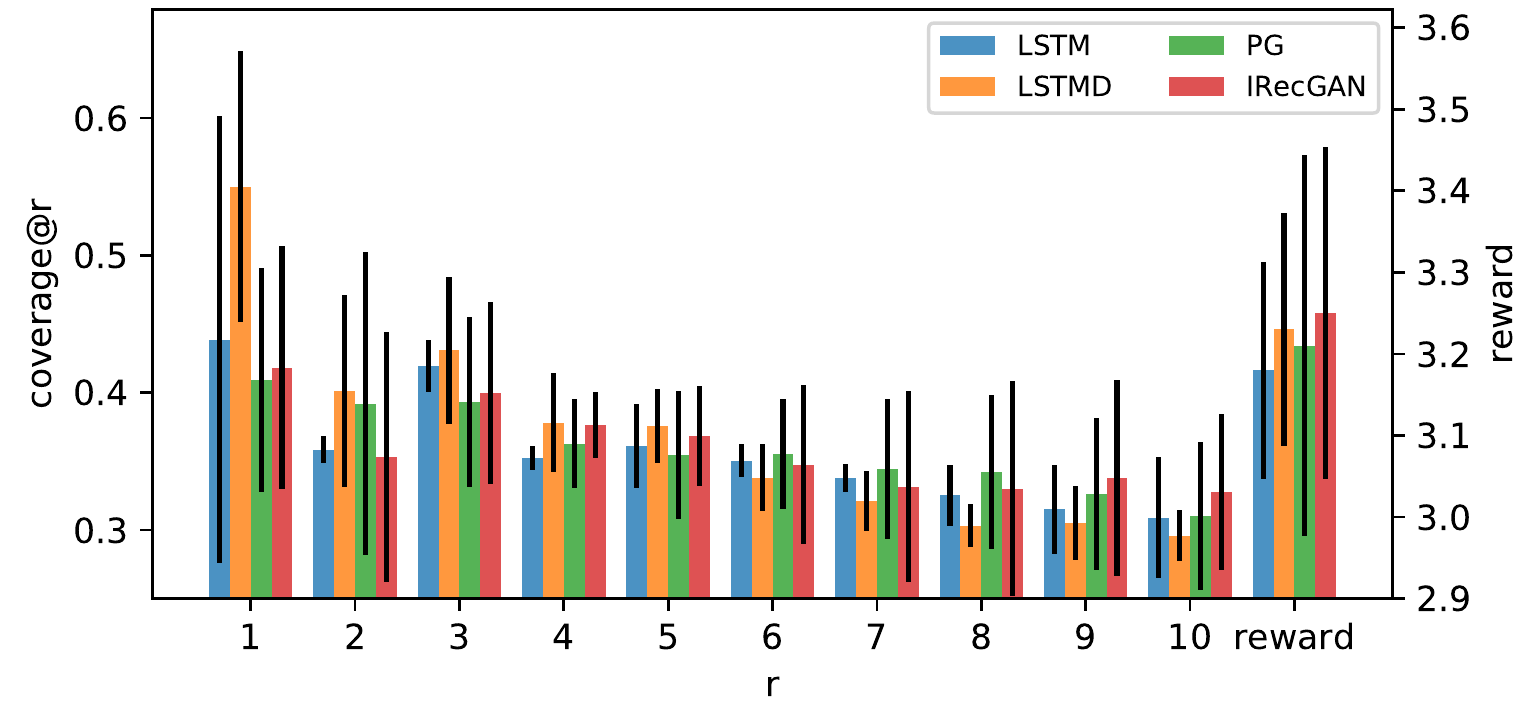}}
\subfigure[$\pi_{\text{random}}(10,000)$]{\includegraphics[width=2.73in]{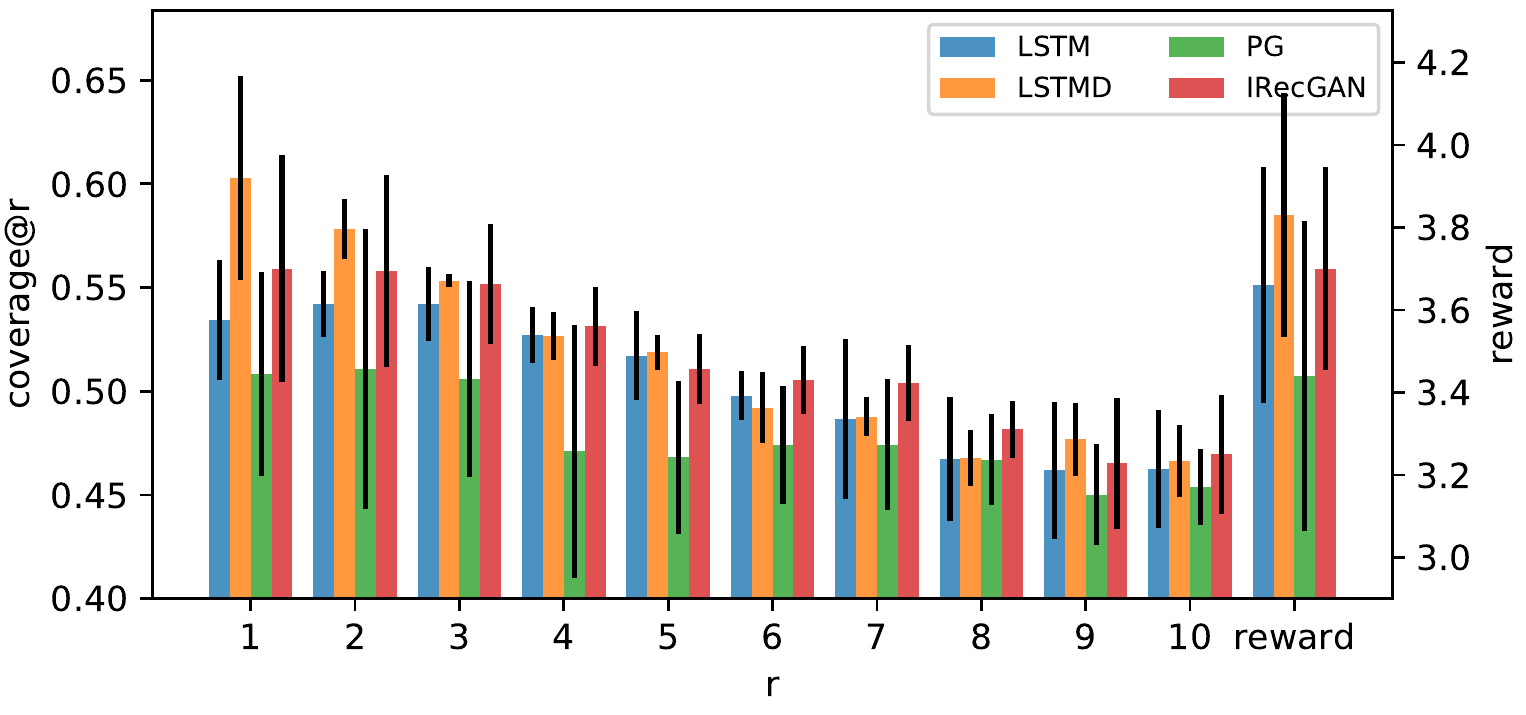}}
\subfigure[$\pi_{\text{max}}(10,000)$]{\includegraphics[width=2.73in]{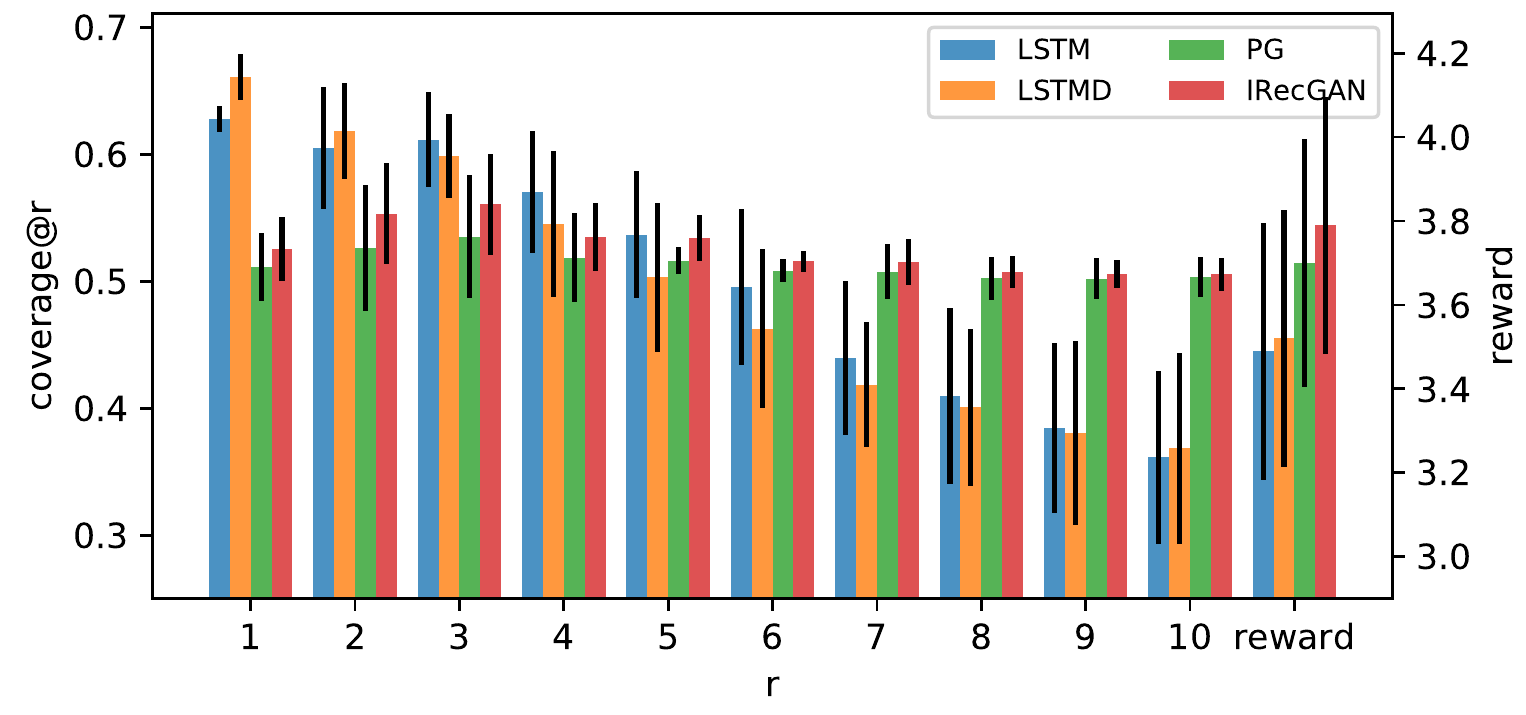}}
\subfigure[$\pi_{\text{mix}}(10,000)$]{\includegraphics[width=2.73in]{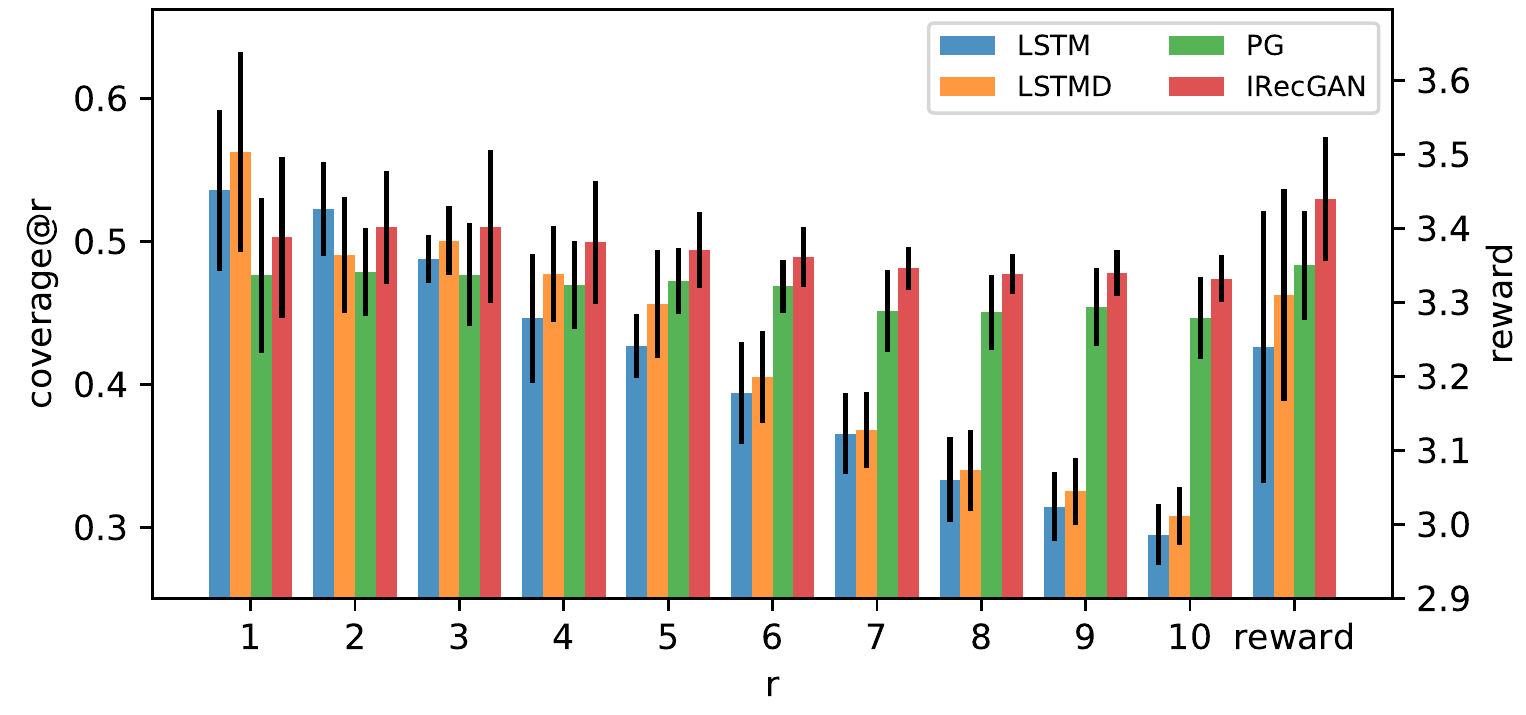}}
\vspace{-3mm}
\caption{Online evaluation results of {coverage@r} and cumulative rewards. }
\label{simu1}
\vspace{-4mm}
\end{figure}

\begin{figure}[t]
\centering
\subfigure{
\includegraphics[width=2.67in]{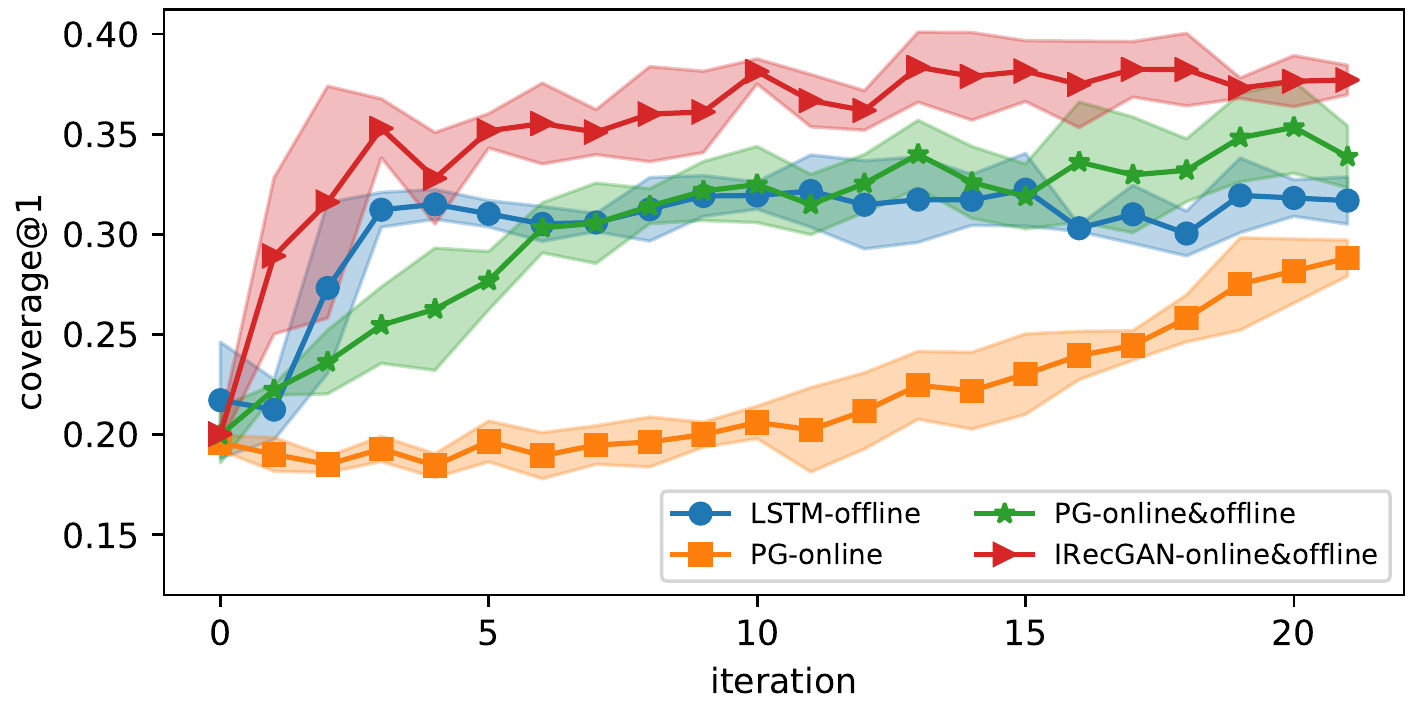}}
\subfigure{
\includegraphics[width=2.67in]{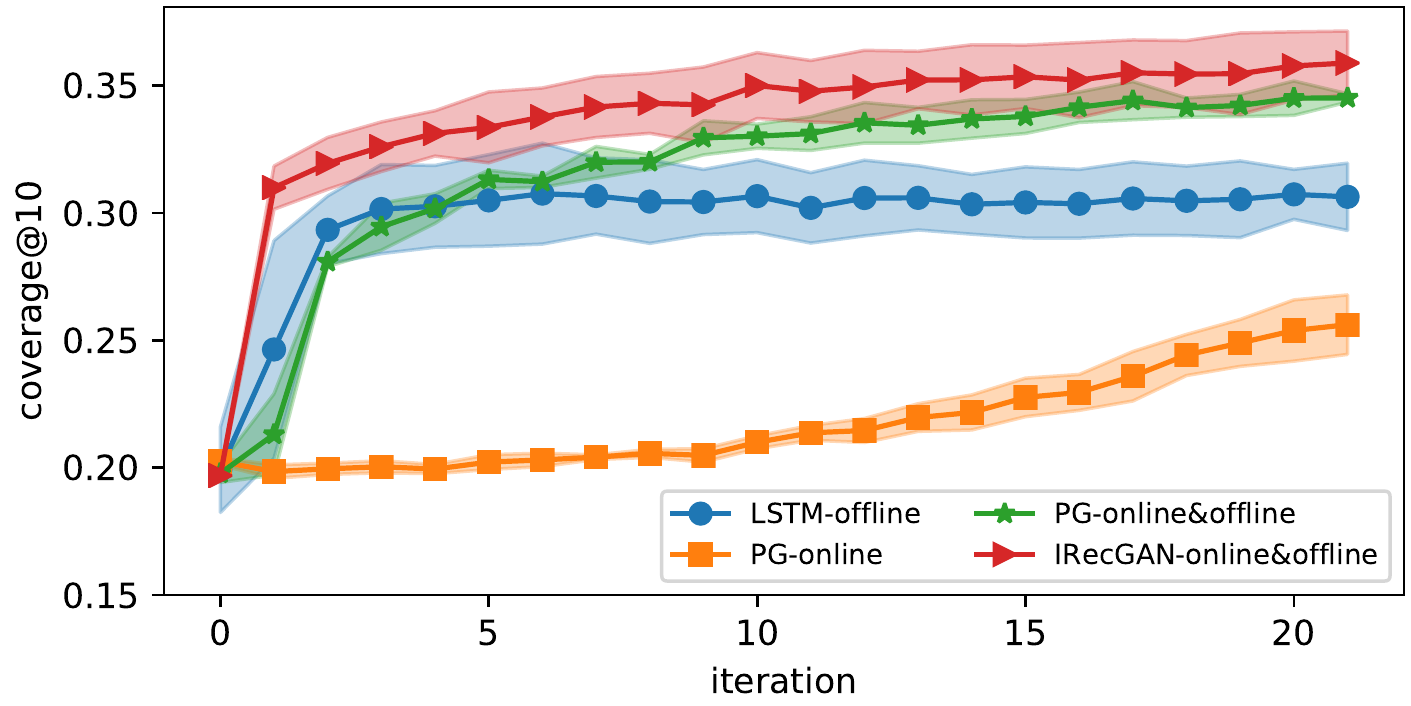}
}
\vspace{-4mm}
\caption{Online learning results of {coverage@1} and {coverage@10}.} 
\label{simu2}
\vspace{-4mm}
\end{figure}

\textbf{Baselines}
We compared our IRecGAN with the following baselines: 1) \textbf{LSTM:} only the user behavior model trained on offline data; 2) \textbf{PG:} only the agent model trained by policy gradient on offline data; 3) \textbf{LSTMD:} the user behavior model in IRecGAN, updated by adversarial training.

\textbf{Experiment Settings}\label{setting}
The hyper-parameters in all models are set as follows: the item embedding dimension is set to 50, the discount factor $\gamma$ in value calculation is set to 0.9, the scale factors $\lambda_r$ and $\lambda_p$ are set to 3 and 1. We use 2-layer LSTM units with 512-dimension hidden states. The ratio of generated training samples and offline data for each training epoch is set to 1:10. We use an RNN based discriminator in all experiments with details provided in the appendix. 

\textbf{Online Evaluation}\label{sim_online_test}
After training our models and baselines on $d_{\text{off}}$, we deploy the learned policy to interact with the simulator for online evaluation. We calculated {coverage@r} to measure the proportion of the true top $r$ relevant items that are ranked in the top $k$ recommended items by a model across all time steps (details in the appendix).
The results of {coverage@r} under different configurations of offline data generation are reported in Figure~\ref{simu1}. 
Under $\pi_{\text{random}}$, coverage@r of all algorithms are relatively low when $r$ is large and the difference in overall performance between behavior and agent models is not very large. This suggests the difficulty of recognizing high reward items under $\pi_{\text{random}}$, because every item has an equal chance to be observed (i.e., full exploration) especially with a small size of  offline data. However, under $\pi_{\text{max}}$ and $\pi_{\text{mix}}$, when the high reward items can be sufficiently learned, user behavior models (LSTM, LSTMD) fail to capture the overall preferred items while agent models (PG, IRecGAN) are stable to the change of $r$. IRecGAN shows its advantage especially under $\pi_{\text{mix}}$, which requires a model to differentiate top relevant items from those with moderate reward. It has close coverage@r to LSTM when $r$ is small and better captures users' overall preferences when user behavior models fail seriously. 
When rewards can not be sufficiently learned (Fig \ref{simu1}(a)), our mechanism can strengthen the influence of truly learned rewards (LSTMD outperforms LSTM when $r$ is small) but may also underestimate some bias. However, when it is feasible to estimate the reward generation (Fig \ref{simu1}(b)(c)(d)), both LSTMD and IRecGAN outperform baselines in coverage@r under the help of generating samples via adversarial training.   

The average cumulative rewards are also reported in the rightmost bars of Figure~\ref{simu1}. They are calculated by generating 1000 sequences with the environment and take the average of their cumulative rewards. IRecGAN has a larger average cumulative reward than other methods under all configurations except $\pi_{\text{random}}$ with 10,000 offline sequences. Under $\pi_{\text{random}}(10,000)$ IRecGAN outperforms PG but not LSTMD. The low cumulative reward of PG under $\pi_{\text{random}}$ indicates that the transition probabilities conditioned on high rewarded items may not be sufficiently learned under the random offline policy.  

 
\noindent\textbf{Online Learning} To evaluate our model's effectiveness in a more practical setting, we execute online and offline learning alternately. Specifically, we separate the learning into two stages: first, the agents can directly interact with the simulator to update their policies, and we only allow them to generate 200 sequences in this stage; then they turn to the offline stage to reuse their generated data for offline learning. We iterate the two stages and record their performance in the online learning stage. We compare with the following baselines: 1) \textbf{PG-online} with only online learning, 2) \textbf{PG-online\&offline} with online learning and reusing the generated data via policy gradient for offline learning, and 3) \textbf{LSTM-offline} with only offline learning. We train all the models from scratch and report the performance of {coverage@1} and {coverage@10} over 20 iterations in Figure~\ref{simu2}. We can observe that LSTM-offline performs worse than other RL methods with offline learning, especially in the later stage, due to its lack of exploration. PG-online improves slowly 
as it does not reuse the generated data. Compared with PG-online\&offline, IRecGAN has better convergence and coverage because of its reduced value estimation bias. We also find that {coverage@10} is harder to improve. The key reason is that as the model identifies the items with high rewards, it tends to recommend them more often. This gives less relevant items less chance to be explored, which is similar to our online evaluation experiments under $\pi_{\text{max}}$ and $\pi_{\text{mix}}$. Our model-based RL training alleviates this bias to a certain extent by generating more training sequences, but it cannot totally alleviate it. This reminds us to focus on explore-exploit trade-off in model-based RL in our future work.

\vspace{-0.1cm}
\subsection{Real-world Data Offline Test}
\vspace{-0.1cm}
\label{offline_test}
We use a large-scale real-world recommendation dataset from CIKM Cup 2016 to evaluate the effectiveness of our proposed solution for offline reranking. Sessions of length 1 or longer than 40 and items that have never been clicked are filtered out. We selected the top 40,000 most popular items into the recommendation candidate set, and randomly selected 65,284/1,718/1,720 sessions for training/validation/testing. The average length of sessions is 2.81/2.80/2.77 respectively; and the ratio of clicks which lead to purchases is 2.31\%/2.46\%/2.45\%.
We followed the same model setting as in our simulation-based study in this experiment. To understand of the effect of different data separation strategies on RL model training and test, we also provide a comparison of performances under different data separation strategies in the appendix.

\textbf{Baselines}
In addition to the baselines we compared in our simulation-based study, we also include the following state-of-the-art solutions for recommendation: 1). \textbf{PGIS:} the agent model estimated with importance sampling on offline data to reduce bias; 2). \textbf{AC:} an LSTM model whose setting is the same as our agent model but trained with actor-critic algorithm~\cite{lillicrap2015continuous} to reduce variance; 3). \textbf{PGU:} the agent model trained using offline and generated data, without adversarial training; 4). \textbf{ACU:} AC model trained with both offline and generated data, without adversarial training.

\textbf{Evaluation Metrics}
All the models were applied to rerank the given recommendation list at each step of testing sessions in offline data. We used Precision@k (P@1 and P@10) to compare different models' recommendation performance, where we define the clicked items as relevant. 
Because the logged recommendation list was not ordered, we cannot assess the logging policy's performance here.

\begin{table}[!ht]
\vspace{-0.1cm}
\caption{Rerank evaluation on real-world dataset with random splitting.}
\vspace{-0.15cm}
\resizebox{\columnwidth}{!}{%
\begin{tabular}{c|cc|cccccc}
\hline
\textbf{Model} & \textbf{LSTM} & \textbf{LSTMD} &\textbf{PG} & \textbf{PGIS} & \textbf{AC} & \textbf{PGU} & \textbf{ACU} &  \textbf{IRecGAN} \\ \hline
\textbf{P@10~(\%)}  & 32.89$\pm$0.50    & 33.42$\pm$0.40   & 33.28$\pm$0.71     & 28.13$\pm$0.45       & 31.93$\pm$0.17     & 34.12$\pm$0.52      & 32.43$\pm$0.22                    & \textbf{35.06}$\pm$0.48    \\
\textbf{P@1~(\%)}   & 8.20$\pm$0.65   & \textbf{8.55}$\pm$0.63     & 6.25$\pm$0.14      & 4.61$\pm$0.73        & 6.54$\pm$0.19      & 6.44$\pm$0.56       & 6.63$\pm$ 0.29           & 6.79$\pm$0.44    \\ \hline
\end{tabular}
}
\label{rerank}
\vspace{-0.3cm}
\end{table}

\paragraph{Results}
The results of the offline rerank evaluation are reported in Table \ref{rerank}. With the help of adversarial training, IRecGAN achieved encouraging P@10 improvement against all baselines. This verifies the effectiveness of our model-based reinforcement learning, especially its adversarial training strategy for utilizing the offline data with reduced bias.
Specifically, 
PGIS did not perform as well as PG partially because of the high variance introduced by importance sampling. 
PGU was able to fit the given data more accurately than PG by learning from the generated data, since there are many items for recommendation and the collected data is limited. However, PGU performed worse than IRecGAN because of the biased user behavior model.
And with the help of the discriminator, IRecGAN reduces the bias in the user behavior model to improve value estimation and policy learning. This is also reflected on its improved user behavior model: LSTMD outperformed LSTM, given both of them are for user behavior modeling.

%% file: conclusion.tex
\vspace{-0.1cm}
\section{Conclusion}
\vspace{-0.1cm}
In this work, we developed a practical solution for utilizing offline data to build a model-based reinforcement learning solution for recommendation. We introduce adversarial training for joint user behavior model learning and policy update. Our theoretical analysis shows our solution's promise in reducing bias; our empirical evaluations in both synthetic and real-world recommendation datasets verify the effectiveness of our solution. Several directions left open in our work, including balancing explore-exploit in policy learning with offline data, incorporating richer structures in user behavior modeling, and exploring the applicability of our solution in other off-policy learning scenarios, such as conversational systems. 

%% file: Supplementary.tex
\appendixpagenumbering
\textbf{\Large Appendix}

\vspace{+0.1cm}
\textbf{\large 1 \hspace{+0.1cm} Details of Discriminator Model}\\
\vspace{-0.3cm}

We adopt an RNN-based discriminator $\mathcal{D}$ for our IRecGAN framework, and model its hidden states by $\mathbf{s}_t^d=h_d(\mathbf{s}_{t-1}^d, \mathbf{e}_{t-1}^d)$, where $\mathbf{s}_t^d$ denotes the hidden states maintained by the discriminator at time $t$ and $\mathbf{e}_{t-1}^d$ is the embedding used in the discriminator side. And we add a multi-layer perceptron which takes the hidden states as input to compute a score through a Sigmoid layer indicating whether the trajectory is likely to be generated by real users when interacting with a recommender as following:
    \begin{align*}
        D(\tau_{0:T}) &= \text{Sigmoid}\big[\frac{1}{T}\sum_{t=0}^{T}\mathbf{e}^d(c^{{\rm max}}_{a_t})^\top \mathbf{e}^d(c_{t})(\text{W}^pr_{t}+\text{b}^p)\big] \label{dis_prob}\\
        c^{{\rm max}}_{a_t}&=\text{argmax}_{c\in a_t}(\mathbf{W}^{d}\mathbf{s}^d_{t}+\mathbf{b}^d)^\top\mathbf{e}({c})
    \end{align*}
    
    where $c^{{\rm max}}_{a_t}$ can be considered as the user's preferred item in the given recommendation list $a_t$, and should be as close to the observed clicks $c_{t}$ as possible for real users. To enable the gradient backpropagation, we use Softmax with a temperature 0.1 to approximate the argmax function. Other hyper parameters are set to the same with the experiment setting depicted in Section~\ref{setting}. The optimization target of $\mathcal{D}$ is formulated as in Eq~(\ref{D_train}).

\vspace{+0.2cm}
\textbf{\large 2 \hspace{+0.1cm} Details of Sampling}

To enable exploration during model training, inspired by the discussion in \cite{chen2019top}, we sample items to get the recommendation list (action) by Eq (\ref{agent_prob}). Moreover, since our model-based RL solution involves the user behavior model (which is estimated together with the agent model) during the sequence generation, we sample users' clicks by their probabilities in the sequence generation of training as well. For testing, the agent's recommendation list contains items with top $k$ probabilities under the learned policy. In the meanwhile, for comparison purpose, we can also use the user behavior model to create a ranking list of items for recommendation purpose. Specifically, in the offline evaluation, user behavior models rerank recorded offline recommendations; and in the simulated online evaluation, user behavior models rerank all items in $A$ (i.e., with given recommendations containing all items) and select the top $k$ for the evaluation of coverage@r. 

\vspace{+0.2cm}
\textbf{\large 3 \hspace{+0.1cm} Algorithm}
\vspace{-0.2cm}
\begin{algorithm}
    \SetAlgoLined
    \KwIn{Offline data; an agent model $\cal A$; a user behavior model $\cal U$; a discriminator $\cal D$.}
    
    Initialize an empty simulated sequences set ${B}^s$ and a real sequences set $B^r$.

    Initialize $\cal U, \cal A$ and $\cal D$ with random parameters.

    Pre-train $\cal U$ by maximizing Eq (\ref{U_pretrain}).

    Pre-train $\cal A$ via the policy gradient of Eq (\ref{all_agent}) using only the offline data.

    $\cal A$ and $\cal U$ simulate $m$ sequences and add them to ${B}^s$. Add $m$ trajectories to real data set $B^r$.

    Pre-train $\cal D$ according to Eq (\ref{D_train}) using ${B}^r$ and ${B}^s$.\\

    \For{$e\leftarrow 1$ \KwTo $epoch$}{
        \For{$\rm r-steps$}{
            Empty ${B}^s$ and then generate $m$ simulated sequences and add to ${B}^s$.

            Compute $q_{\cal D}(\tau_{0:t})$ at each step $t$ by Eq (\ref{roll-out}).
            
            Extract $\left\lfloor {\frac{{{\lambda _2}}}{{{\lambda _1}}}m} \right\rfloor$ sequences into $B^r$.
                
            Update $\cal U$ via the policy gradient of Eq (\ref{seq_usr}) with $B=[B^s, B^r]$.
            
            Update $\cal A$ via the policy gradient of Eq (\ref{all_agent}) with $B=[B^s, B^r]$.
            
        }
        \For{$\rm d-steps$}{
            Empty ${B}^s$, then generate $m$ simulated sequences by current $\cal U$, $\cal A$ and add to ${B}^s$. 
            
            Empty ${B}^r$ and add $m$ sequences from the offline data.
           
            Update $\cal D$ according to Eq (\ref{D_train}) for $i$ epochs using ${B}^r$ and ${B}^s$.
        }
    }
    \caption{IRecGAN}
\end{algorithm}

\textbf{\large 4 \hspace{+0.1cm} The Weight of Sequence Generation Score}

A weight $w$ can be applied to the sequence generation score $q_D$ for purpose of rescaling the generated rewards. In this paper, we set $w=1$ and got the expected value estimation in Section 6. In this setting, when the agent's policy is the same as that of the offline data and the user behavior model is unbiased, which means $P_{{\pi _{{\Theta_a}}}}(\tau _{0:t}^{}) = P_{{{{g}}}}(\tau _{0:t}^{}) = P_{{{{data}}}}(\tau _{0:t}^{})$ and $\Delta=0$, the value estimation is biased. By setting $w=2$, the expected value estimation ${\mathbb E}_{\tau  \sim {\pi _{_{{\Theta_a}}}}}[V_a]$ turns out to be:
\small
\[V_a^{{\pi _{{\Theta _a}}}} + \sum\limits_{t = 0}^T {{\mathbb E_{{\tau _{0:t}}\sim{\pi _{{\Theta _a}}}}}\frac{{{2}{\lambda _1}}}{{2 - ({\delta _1} + {\delta _2})}}} \Delta  + \sum\limits_{t = 0}^T {{\mathbb E_{{\tau _{0:t}}\sim data}}{\lambda _2}{\delta _2}{r_t}} ({\tau _{0:t}}) + \sum\limits_{t = 0}^T {{\mathbb E_{{\tau _{0:t}}\sim{\pi _{{\Theta _a}}}}}} \frac{{({\delta _1} + {\delta _2}){\lambda _1}}}{{2 - ({\delta _1} + {\delta _2})}}{r_t}({\tau _{0:t}}).\]
\normalsize
This value estimation is unbiased when $P_{{\pi _{{\Theta_a}}}}(\tau _{0:t}^{}) = P_{{{{g}}}}(\tau _{0:t}^{}) = P_{{{{data}}}}(\tau _{0:t}^{})$ and $\Delta=0$.

However, when the user behavior model is biased, amplifying the sequence generation score with $w > 1$ will also amplify the bias. Moreover, it will over-penalize the generated sequences which are not very similar to the offline data (with relatively low $q_{\cal D}$). Although our method encourages the agent to consider users' immediate clicks when making recommendations, it does not require the overall recommendations to be similar to those of the offline data. And we also do not want the agent's recommendations to be exactly the same as the recorded ones, since our goal is to improve the offline policy. In this case, over-penalizing some generated sequences is harmful. Because of the reasons above, we directly use $q_{\cal D}$ in our paper. But we admit that the weight $w$ can be set to different values under specific cases for value estimation.

\vspace{+0.2cm}
\textbf{\large 5 \hspace{+0.1cm} Details about the Coverage Metric}

In our simulated environment, the selection of a click directly relates to its reward, which also influences the length of the sequence. In this case, whether the model (the user behavior model or the agent) can capture real  reward of items at each time will highly affect its performance in both behavior prediction and recommendation. As indicated in Section \ref{sim_online_test}, we use coverage@r to measure whether a model can capture items with high rewards (most relevant items) under corresponding states. Denote the top $r$ relevant items at time $t$ as $C^r_t$, the top $k$ recommendations given by the model as $A^k_t$. 
We regard the $k$ items with the highest prediction scores from a recommendation algorithm (a user behavior model can also be treated as a recommendation algorithm when the click candidates are from the whole item set) as its recommendations given the whole item set as its candidates. Then the coverage@r can be calculated by 
\[\text{coverage@r}=\frac{{\sum\nolimits_{t = 0}^T {\left| {C_t^r \cap A_t^k} \right|} }}{{T\times r}}.\]
When $r$ is small, it requires models to capture the most relevant items to get a high coverage@r. When $r$ becomes larger, models which can capture overall high reward items are likely to get high coverage@r. For example, an evaluation result with high coverage@1 and low coverage@2 indicates the algorithm handles the highest reward item in the ground-truth better than the second item. 

To the behavior model, since the environment's next clicks are sampled according to the items' conditional rewards with respect to the state, a model's coverage@r performance directly relates to its performance of the behavior prediction (especially when $r$ is small). To the agent, since it aims to maximize the cumulative rewards, including items with relatively high immediate rewards will ensure users' satisfaction and the model's immediate gain at each time step. Moreover, since items with high rewards also have high success probabilities in the Bernoulli experiment, ensuring users' clicking of high reward items encourages the continuation of sequences, which also improves the accumulation of rewards. Because of these, the agent's coverage@r performance is highly related to the actual cumulative rewards it can get in our simulated environment. However, different from the behavior model, because the cumulative rewards an agent can get also relate to the state transitions conditioned on the clicked items, a performing agent should not always recommend items with the highest rewards. In this case, we also provide an evaluation of cumulative rewards in the results.

\vspace{+0.2cm}
\textbf{\large 6 \hspace{+0.1cm} Correction of Figure~\ref{simu1} and Figure~\ref{simu2}}

Compared to the original version, we have corrected Figure~\ref{simu1} and Figure~\ref{simu2} about the results of simulated experiments. This is because of an implementation mismatch when computing the simulated environment. Specifically, in the previous implementation the next click under the state $s_i$ is re-selected by $\arg {\max _{{a_i}}}r({a_i} \in A_t^k|{s_i})$ instead of $a_j$, after the success of the Bernoulli experiment with the probability $r(a_j|s_i)$. This leads to a situation in which all methods are hard to estimate rewards of items with relatively low ground-truth rewards in $s_i$, no matter under $\pi_\text{random}$ or $\pi_\text{max}$. This leads to the performance drop of coverage@r with the increase of $r$. After the correction, the updated results and their corresponding analysis are shown in Section \ref{sim_online_test}. 

\vspace{+0.2cm}\textbf{\large 7 \hspace{+0.1cm}Offline Test with Different Data Separations}

In real-world data offline evaluation of Section~\ref{offline_test}, since we do not know the logging policy of the offline recommendation, the true distribution of data appearing under the offline recommendation policy can only be inferred by the observations. However, because the problem space of our offline dataset is large, it is hard to sufficiently reveal the true data distribution with limited offline data. In this case, using different data separation strategies may lead to different data distributions for both training and testing, which may cause different performance of models as indicated in our simulated online evaluation. To provide a more comprehensive evaluation, we randomly split the dataset for training/validation/testing in Section~\ref{offline_test}. We adopted P@1 and P@10 to compare different models' performance. And both the metrics were calculated only on the timesteps with a recommendation list including more than 10 candidate items. Moreover, we conducted the offline evaluation experiments three times by varying the random seed to get the confidence interval for each algorithm. 

To compare results under different data separation strategies, we evaluated models when splitting the dataset in the order of session ID or time, as shown in Table \ref{rerank_2} and \ref{rerank_3}, respectively. Specifically, we ordered the whole dataset by session ID or time, and used 65,284/1,718/1,820 sessions for training/validation/testing. When split data by session ID, the average length of training/validation/testing sessions was 2.84/2.15/2.09, the ratio of clicks that lead to purchases was 2.32\%/2.08\%/2.36\%. When split data by time, the average length of training/validation/testing sessions was 2.81/2.80/2.75 and the ratio of clicks leading to purchases was 2.33\%/2.21\%/2.05\%. And to provide more insights about performance of different algorithms, we also included P@1~(all) that measures Precision@1 on all the timesteps (with more than one recommendation candidate) for each model as a metric.

\begin{table}[!ht]
\caption{Rerank evaluation on real-world recommendation dataset when split by session ID.}
\vspace{-0.1cm}
\resizebox{\columnwidth}{!}{%
\begin{tabular}{c|cc|cccccc}

\hline
\textbf{Model} & \textbf{LSTM} & \textbf{LSTMD} &\textbf{PG} & \textbf{PGIS} & \textbf{AC} & \textbf{PGU} & \textbf{ACU} &  \textbf{IRecGAN} \\ \hline
\textbf{P@10~(\%)}  & 
28.79$\pm$0.44& 31.98$\pm$0.64 & 32.44$\pm$1.16     & 30.72$\pm$0.37       & 29.26$\pm$0.79     & 30.33$\pm$0.47 & 28.53$\pm$0.35     & \textbf{33.45}$\pm$0.71    \\
\textbf{P@1~(\%, all)}   & 
9.64$\pm$0.38& \textbf{11.26}$\pm$0.34& 8.40$\pm$0.18& 7.67$\pm$0.31& 7.33$\pm$0.41& 8.27$\pm$0.44& 7.08$\pm$0.32& 9.78$\pm$0.37
\\
\textbf{P@1~(\%)}   & 
9.68$\pm$0.29&\textbf{11.06}$\pm$0.23&6.83$\pm$0.38
&6.09$\pm$0.19&6.11$\pm$0.18&6.67$\pm$0.51&5.86$\pm$0.26& 7.84$\pm$0.25
\\ \hline
\end{tabular}
}
\label{rerank_2}
\vspace{-0.3cm}
\end{table}

\begin{table}[!ht]
\caption{Rerank evaluation on real-world recommendation dataset when split by time.}
\vspace{-0.1cm}
\resizebox{\columnwidth}{!}{%
\begin{tabular}{c|cc|cccccc}

\hline
\textbf{Model} & \textbf{LSTM} & \textbf{LSTMD} &\textbf{PG} & \textbf{PGIS} & \textbf{AC} & \textbf{PGU} & \textbf{ACU} &  \textbf{IRecGAN} \\ \hline
\textbf{P@10~(\%)}  & 
27.95$\pm$0.34& 29.85$\pm$0.18 & 29.13$\pm$0.18 & 27.85$\pm$0.15&25.37$\pm$0.49&29.45$\pm$0.37&26.51$\pm$0.67&\textbf{30.07}$\pm$0.15\\
\textbf{P@1~(\%, all)}   & 7.94$\pm$0.10&\textbf{8.27}$\pm$0.14&6.07$\pm$0.15&6.91$\pm$0.11&4.08$\pm$0.12&6.58$\pm$0.18&4.84$\pm$0.23&7.08$\pm$0.25\\
\textbf{P@1~(\%)}   & 
7.67$\pm$0.12&\textbf{7.90}$\pm$0.14&4.65$\pm$0.25&5.40$\pm$0.13&4.16$\pm$0.15&5.19$\pm$0.27&4.89$\pm$0.21&5.81$\pm$0.18\\\hline
\end{tabular}
}
\label{rerank_3}
\end{table}
We observed that the results had a considerable difference compared with random data separation when we split the data by session ID or time, which validated the influence of data separation. However, the overall conclusions in the comparison among our methods (LSTMD, IRecGAN) and baselines remained consistent. Because of their different training purposes where user behavior models (LSTM, LSTMD) were trained only for click prediction, LSTM and LSTMD performed better than the RL agents in P@1. And the RL agents (IRecGAN and other RL baselines) had advantages in capturing users' overall interests, which led to better P@10 results. 

Although we observed different performance of baselines under different data separation strategies, using our additional sample generation mechanism with adversarial training and under all strategies, LSTMD outperformed LSTM and IRecGAN outperformed all other RL-based methods in both P@1 and P@10. These results showed that 1) the adversarial training solution we proposed in this paper helped to improve the user behavior model. 2) The sequence generation reward for the RL agent helped it better capture users' immediate behaviors. 
3) The proposed solution helped the agent to better capture users' overall interests.

By comparing the P@1 and P@1 (all), we observed that the differences between these two metrics in user behavior models (LSTM and LSTMD) were smaller than those between most RL agents. More specifically, most of RL agents performed better under the P@1 (all) metric than P@1, where the former included evaluations with less than 10 ranking candidates. We conjecture that the key reason is that user behavior models are only optimized for  click prediction, while a RL agent needs to balance both the next click and future clicks via the learnt state transition. 
When the number of recommendation candidates to re-rank is small, there is more chance that an agent ranks the next click on top, which leads to a better P@1 (all).


%% file: neurips_2019.bbl
\begin{thebibliography}{38}
\providecommand{\natexlab}[1]{#1}
\providecommand{\url}[1]{\texttt{#1}}
\expandafter\ifx\csname urlstyle\endcsname\relax
  \providecommand{\doi}[1]{doi: #1}\else
  \providecommand{\doi}{doi: \begingroup \urlstyle{rm}\Url}\fi

\bibitem[Achiam et~al.(2017)Achiam, Held, Tamar, and
  Abbeel]{achiam2017constrained}
Joshua Achiam, David Held, Aviv Tamar, and Pieter Abbeel.
\newblock Constrained policy optimization.
\newblock In \emph{Proceedings of the 34th International Conference on Machine
  Learning-Volume 70}, pages 22--31. JMLR. org, 2017.

\bibitem[Chen et~al.(2019{\natexlab{a}})Chen, Beutel, Covington, Jain,
  Belletti, and Chi]{chen2019top}
Minmin Chen, Alex Beutel, Paul Covington, Sagar Jain, Francois Belletti, and
  Ed~H Chi.
\newblock Top-k off-policy correction for a reinforce recommender system.
\newblock In \emph{Proceedings of the Twelfth ACM International Conference on
  Web Search and Data Mining}, pages 456--464. ACM, 2019{\natexlab{a}}.

\bibitem[Chen et~al.(2019{\natexlab{b}})Chen, Li, Li, Jiang, Qi, and
  Song]{pmlr-v97-chen19f}
Xinshi Chen, Shuang Li, Hui Li, Shaohua Jiang, Yuan Qi, and Le~Song.
\newblock Generative adversarial user model for reinforcement learning based
  recommendation system.
\newblock In \emph{Proceedings of the 36th International Conference on Machine
  Learning}, volume~97, pages 1052--1061, 2019{\natexlab{b}}.

\bibitem[Chung et~al.(2014)Chung, Gulcehre, Cho, and
  Bengio]{chung2014empirical}
Junyoung Chung, Caglar Gulcehre, KyungHyun Cho, and Yoshua Bengio.
\newblock Empirical evaluation of gated recurrent neural networks on sequence
  modeling.
\newblock \emph{arXiv preprint arXiv:1412.3555}, 2014.

\bibitem[Deisenroth and Rasmussen(2011)]{deisenroth2011pilco}
Marc Deisenroth and Carl~E Rasmussen.
\newblock Pilco: A model-based and data-efficient approach to policy search.
\newblock In \emph{Proceedings of the 28th International Conference on machine
  learning (ICML-11)}, pages 465--472, 2011.

\bibitem[Deisenroth et~al.(2011)Deisenroth, Rasmussen, and
  Fox]{deisenroth2011learning}
Marc~Peter Deisenroth, Carl~Edward Rasmussen, and Dieter Fox.
\newblock Learning to control a low-cost manipulator using data-efficient
  reinforcement learning.
\newblock 2011.

\bibitem[Deisenroth et~al.(2013)Deisenroth, Neumann, Peters,
  et~al.]{deisenroth2013survey}
Marc~Peter Deisenroth, Gerhard Neumann, Jan Peters, et~al.
\newblock A survey on policy search for robotics.
\newblock \emph{Foundations and Trends{\textregistered} in Robotics},
  2\penalty0 (1--2):\penalty0 1--142, 2013.

\bibitem[Gilotte et~al.(2018)Gilotte, Calauz{\`e}nes, Nedelec, Abraham, and
  Doll{\'e}]{gilotte2018offline}
Alexandre Gilotte, Cl{\'e}ment Calauz{\`e}nes, Thomas Nedelec, Alexandre
  Abraham, and Simon Doll{\'e}.
\newblock Offline a/b testing for recommender systems.
\newblock In \emph{Proceedings of the Eleventh ACM International Conference on
  Web Search and Data Mining}, pages 198--206. ACM, 2018.

\bibitem[Goodfellow et~al.(2014)Goodfellow, Pouget-Abadie, Mirza, Xu,
  Warde-Farley, Ozair, Courville, and Bengio]{goodfellow2014generative}
Ian Goodfellow, Jean Pouget-Abadie, Mehdi Mirza, Bing Xu, David Warde-Farley,
  Sherjil Ozair, Aaron Courville, and Yoshua Bengio.
\newblock Generative adversarial nets.
\newblock In \emph{Advances in neural information processing systems}, pages
  2672--2680, 2014.

\bibitem[Gu et~al.(2016)Gu, Lillicrap, Sutskever, and Levine]{gu2016continuous}
Shixiang Gu, Timothy Lillicrap, Ilya Sutskever, and Sergey Levine.
\newblock Continuous deep q-learning with model-based acceleration.
\newblock In \emph{International Conference on Machine Learning}, pages
  2829--2838, 2016.

\bibitem[He et~al.(2016)He, Zhang, Kan, and Chua]{he2016fast}
Xiangnan He, Hanwang Zhang, Min-Yen Kan, and Tat-Seng Chua.
\newblock Fast matrix factorization for online recommendation with implicit
  feedback.
\newblock In \emph{Proceedings of the 39th International ACM SIGIR conference
  on Research and Development in Information Retrieval}, pages 549--558. ACM,
  2016.

\bibitem[Hochreiter and Schmidhuber(1997)]{hochreiter1997long}
Sepp Hochreiter and J{\"u}rgen Schmidhuber.
\newblock Long short-term memory.
\newblock \emph{Neural computation}, 9\penalty0 (8):\penalty0 1735--1780, 1997.

\bibitem[Koren et~al.(2009)Koren, Bell, and Volinsky]{koren2009matrix}
Yehuda Koren, Robert Bell, and Chris Volinsky.
\newblock Matrix factorization techniques for recommender systems.
\newblock \emph{Computer}, \penalty0 (8):\penalty0 30--37, 2009.

\bibitem[Learning(1998)]{learning1998introduction}
Reinforcement Learning.
\newblock An introduction, richard s. sutton and andrew g. barto, 1998.

\bibitem[Liebman et~al.(2015)Liebman, Saar-Tsechansky, and
  Stone]{liebman2015dj}
Elad Liebman, Maytal Saar-Tsechansky, and Peter Stone.
\newblock Dj-mc: A reinforcement-learning agent for music playlist
  recommendation.
\newblock In \emph{Proceedings of the 2015 International Conference on
  Autonomous Agents and Multiagent Systems}, pages 591--599. International
  Foundation for Autonomous Agents and Multiagent Systems, 2015.

\bibitem[Lillicrap et~al.(2015)Lillicrap, Hunt, Pritzel, Heess, Erez, Tassa,
  Silver, and Wierstra]{lillicrap2015continuous}
Timothy~P Lillicrap, Jonathan~J Hunt, Alexander Pritzel, Nicolas Heess, Tom
  Erez, Yuval Tassa, David Silver, and Daan Wierstra.
\newblock Continuous control with deep reinforcement learning.
\newblock \emph{arXiv preprint arXiv:1509.02971}, 2015.

\bibitem[Lu and Yang(2016)]{lu2016partially}
Zhongqi Lu and Qiang Yang.
\newblock Partially observable markov decision process for recommender systems.
\newblock \emph{arXiv preprint arXiv:1608.07793}, 2016.

\bibitem[Meger et~al.(2015)Meger, Higuera, Xu, Giguere, and
  Dudek]{meger2015learning}
David Meger, Juan Camilo~Gamboa Higuera, Anqi Xu, Philippe Giguere, and Gregory
  Dudek.
\newblock Learning legged swimming gaits from experience.
\newblock In \emph{2015 IEEE International Conference on Robotics and
  Automation (ICRA)}, pages 2332--2338. IEEE, 2015.

\bibitem[Mnih et~al.(2013)Mnih, Kavukcuoglu, Silver, Graves, Antonoglou,
  Wierstra, and Riedmiller]{mnih2013playing}
Volodymyr Mnih, Koray Kavukcuoglu, David Silver, Alex Graves, Ioannis
  Antonoglou, Daan Wierstra, and Martin Riedmiller.
\newblock Playing atari with deep reinforcement learning.
\newblock \emph{arXiv preprint arXiv:1312.5602}, 2013.

\bibitem[Mnih et~al.(2015)Mnih, Kavukcuoglu, Silver, Rusu, Veness, Bellemare,
  Graves, Riedmiller, Fidjeland, Ostrovski, et~al.]{mnih2015human}
Volodymyr Mnih, Koray Kavukcuoglu, David Silver, Andrei~A Rusu, Joel Veness,
  Marc~G Bellemare, Alex Graves, Martin Riedmiller, Andreas~K Fidjeland, Georg
  Ostrovski, et~al.
\newblock Human-level control through deep reinforcement learning.
\newblock \emph{Nature}, 518\penalty0 (7540):\penalty0 529, 2015.

\bibitem[Morimoto and Atkeson(2003)]{morimoto2003minimax}
Jun Morimoto and Christopher~G Atkeson.
\newblock Minimax differential dynamic programming: An application to robust
  biped walking.
\newblock In \emph{Advances in neural information processing systems}, pages
  1563--1570, 2003.

\bibitem[Munos et~al.(2016)Munos, Stepleton, Harutyunyan, and
  Bellemare]{munos2016safe}
R{\'e}mi Munos, Tom Stepleton, Anna Harutyunyan, and Marc Bellemare.
\newblock Safe and efficient off-policy reinforcement learning.
\newblock In \emph{Advances in Neural Information Processing Systems}, pages
  1054--1062, 2016.

\bibitem[Oh et~al.(2017)Oh, Singh, and Lee]{oh2017value}
Junhyuk Oh, Satinder Singh, and Honglak Lee.
\newblock Value prediction network.
\newblock In \emph{Advances in Neural Information Processing Systems}, pages
  6118--6128, 2017.

\bibitem[Peng et~al.(2018)Peng, Li, Gao, Liu, Wong, and Su]{peng2018deep}
Baolin Peng, Xiujun Li, Jianfeng Gao, Jingjing Liu, Kam-Fai Wong, and Shang-Yu
  Su.
\newblock Deep dyna-q: Integrating planning for task-completion dialogue policy
  learning.
\newblock \emph{arXiv preprint arXiv:1801.06176}, 2018.

\bibitem[Precup(2000)]{precup2000eligibility}
Doina Precup.
\newblock Eligibility traces for off-policy policy evaluation.
\newblock \emph{Computer Science Department Faculty Publication Series},
  page~80, 2000.

\bibitem[Precup et~al.(2001)Precup, Sutton, and Dasgupta]{precup2001off}
Doina Precup, Richard~S Sutton, and Sanjoy Dasgupta.
\newblock Off-policy temporal-difference learning with function approximation.
\newblock In \emph{ICML}, pages 417--424, 2001.

\bibitem[Schulman et~al.(2015)Schulman, Levine, Abbeel, Jordan, and
  Moritz]{schulman2015trust}
John Schulman, Sergey Levine, Pieter Abbeel, Michael Jordan, and Philipp
  Moritz.
\newblock Trust region policy optimization.
\newblock In \emph{International Conference on Machine Learning}, pages
  1889--1897, 2015.

\bibitem[Shani et~al.(2005)Shani, Heckerman, and Brafman]{shani2005mdp}
Guy Shani, David Heckerman, and Ronen~I Brafman.
\newblock An mdp-based recommender system.
\newblock \emph{Journal of Machine Learning Research}, 6\penalty0
  (Sep):\penalty0 1265--1295, 2005.

\bibitem[Sutton(1990)]{sutton1990integrated}
Richard~S Sutton.
\newblock Integrated architectures for learning, planning, and reacting based
  on approximating dynamic programming.
\newblock In \emph{Machine Learning Proceedings 1990}, pages 216--224.
  Elsevier, 1990.

\bibitem[Sutton et~al.(2000)Sutton, McAllester, Singh, and
  Mansour]{sutton2000policy}
Richard~S Sutton, David~A McAllester, Satinder~P Singh, and Yishay Mansour.
\newblock Policy gradient methods for reinforcement learning with function
  approximation.
\newblock In \emph{Advances in neural information processing systems}, pages
  1057--1063, 2000.

\bibitem[Swaminathan and Joachims(2015{\natexlab{a}})]{swaminathan2015batch}
Adith Swaminathan and Thorsten Joachims.
\newblock Batch learning from logged bandit feedback through counterfactual
  risk minimization.
\newblock \emph{Journal of Machine Learning Research}, 16\penalty0
  (1):\penalty0 1731--1755, 2015{\natexlab{a}}.

\bibitem[Swaminathan and Joachims(2015{\natexlab{b}})]{swaminathan2015self}
Adith Swaminathan and Thorsten Joachims.
\newblock The self-normalized estimator for counterfactual learning.
\newblock In \emph{advances in neural information processing systems}, pages
  3231--3239, 2015{\natexlab{b}}.

\bibitem[Thomas and Brunskill(2016)]{thomas2016data}
Philip Thomas and Emma Brunskill.
\newblock Data-efficient off-policy policy evaluation for reinforcement
  learning.
\newblock In \emph{International Conference on Machine Learning}, pages
  2139--2148, 2016.

\bibitem[Williams(1992)]{williams1992simple}
Ronald~J Williams.
\newblock Simple statistical gradient-following algorithms for connectionist
  reinforcement learning.
\newblock \emph{Machine learning}, 8\penalty0 (3-4):\penalty0 229--256, 1992.

\bibitem[Wu et~al.(2017)Wu, Wang, Hong, and Shi]{wu2017returning}
Qingyun Wu, Hongning Wang, Liangjie Hong, and Yue Shi.
\newblock Returning is believing: Optimizing long-term user engagement in
  recommender systems.
\newblock In \emph{Proceedings of the 2017 ACM on Conference on Information and
  Knowledge Management}, pages 1927--1936. ACM, 2017.

\bibitem[Yu et~al.(2017)Yu, Zhang, Wang, and Yu]{yu2017seqgan}
Lantao Yu, Weinan Zhang, Jun Wang, and Yong Yu.
\newblock Seqgan: Sequence generative adversarial nets with policy gradient.
\newblock In \emph{Thirty-First AAAI Conference on Artificial Intelligence},
  2017.

\bibitem[Zhao et~al.(2019)Zhao, Xia, Zhao, Yin, and Tang]{zhao2019model}
Xiangyu Zhao, Long Xia, Yihong Zhao, Dawei Yin, and Jiliang Tang.
\newblock Model-based reinforcement learning for whole-chain recommendations.
\newblock \emph{arXiv preprint arXiv:1902.03987}, 2019.

\bibitem[Zheng et~al.(2018)Zheng, Zhang, Zheng, Xiang, Yuan, Xie, and
  Li]{zheng2018drn}
Guanjie Zheng, Fuzheng Zhang, Zihan Zheng, Yang Xiang, Nicholas~Jing Yuan, Xing
  Xie, and Zhenhui Li.
\newblock Drn: A deep reinforcement learning framework for news recommendation.
\newblock In \emph{Proceedings of the 2018 World Wide Web Conference on World
  Wide Web}, pages 167--176. International World Wide Web Conferences Steering
  Committee, 2018.

\end{thebibliography}
